\pdfoutput=1

\documentclass[11pt]{article}

\usepackage[]{acl}

\usepackage{times}
\usepackage{latexsym}
\usepackage{amssymb}

\usepackage{amsmath}

\usepackage{algorithm}
\usepackage{algpseudocode}

\usepackage[T1]{fontenc}

\usepackage[utf8]{inputenc}

\usepackage{microtype}

\usepackage{inconsolata}

\usepackage{tcolorbox}
\usepackage{color}

\usepackage{booktabs} %
\usepackage{bbding}

\usepackage{subcaption}

\usepackage{enumitem}
\usepackage{hyperref}
\usepackage{pifont}

\usepackage[colorinlistoftodos]{todonotes}

\newcommand{\xmark}{\textcolor{red}{\textbf{\ding{55}}}}
\newcommand{\cmark}{\textcolor{darkgreen}{\textbf{\ding{51}}}}

\algrenewcommand\algorithmicthen{}
\algrenewcommand\algorithmicdo{}
\algrenewcommand\algorithmicrequire{\textbf{Input:}}
\algrenewcommand\algorithmicensure{\textbf{Output:}}

\definecolor{darkgreen}{HTML}{228B22}
\definecolor{nblue}{HTML}{377eb8}

\definecolor{ggreen}{HTML}{38761d}
\definecolor{oorange}{HTML}{d95f02}
\definecolor{bblue}{HTML}{7570b3}
\definecolor{ppurple}{HTML}{e37fbb}

\definecolor{lgreen}{HTML}{9CD24A}
\definecolor{yyellow}{HTML}{FFD52D}
\definecolor{ggold}{HTML}{E1BC89}
\definecolor{ggray}{HTML}{AAAAAA}

\definecolor{crimson}{HTML}{cc0000}

\newtcolorbox[list inside=prompt,auto counter,number within=section]{prompt}[1][]{
    fontupper=\ttfamily\footnotesize,
    boxsep=5pt,
    left=0pt,
    right=0pt,
    top=0pt,
    bottom=0pt,
    boxrule=1pt,
    #1,
}

\title{Retrieval Helps or Hurts? A Deeper Dive into the Efficacy of \\Retrieval Augmentation to Language Models}

\author{
Seiji Maekawa \quad Hayate Iso \quad Sairam Gurajada \quad Nikita Bhutani \\
Megagon Labs \\
\texttt{\{seiji,hayate,sairam,nikita\}@megagon.ai}
}

\begin{document}
\maketitle

\begin{abstract}

While large language models (LMs) demonstrate remarkable performance, they encounter challenges in providing accurate responses when queried for information beyond their pre-trained memorization. Although augmenting them with relevant external information can mitigate these issues, failure to consider the necessity of retrieval may adversely affect overall performance. Previous research has primarily focused on examining how entities influence retrieval models and knowledge recall in LMs, leaving other aspects relatively unexplored. In this work, our goal is to offer a more detailed, fact-centric analysis by exploring the effects of combinations of entities and relations. To facilitate this, we construct a new question answering (QA) dataset called \textsc{WiTQA} (Wikipedia Triple Question Answers). This dataset includes questions about entities and relations of various popularity levels, each accompanied by a supporting passage. Our extensive experiments with diverse LMs and retrievers reveal when retrieval does not consistently enhance LMs from the viewpoints of fact-centric popularity. Confirming earlier findings, we observe that larger LMs excel in recalling popular facts. However, they notably encounter difficulty with infrequent entity-relation pairs compared to retrievers. Interestingly, they can effectively retain popular relations of less common entities. We demonstrate the efficacy of our finer-grained metric and insights through an adaptive retrieval system that selectively employs retrieval and recall based on the frequencies of entities and relations in the question.\footnote{The code and data are available at {\url{https://github.com/megagonlabs/witqa}}.}

\end{abstract}

\section{Introduction}
\label{sec:intro}

{Large language models (LMs) ~\cite{brown2020gpt3,openai2023gpt4} have exhibited impressive capabilities owing to their ability to retrieve knowledge memorized during pre-training ~\cite{sanh2022multitask,wei2022finetuned,ouyang2022training}. However, despite the increase in model size and complexity, LMs remain susceptible to factual inaccuracies in knowledge-intensive tasks such as open domain question answering and natural language generation, which demand access to a broader spectrum of information ~\cite{petroni-etal-2021-kilt,chen-etal-2017-reading,lin2021truthfulqa}. Retrieval-Augmented Language Models (RALMs) \cite{pmlr-v119-guu20a,lewis2020retrieval,izacard-grave-2021-leveraging} have emerged as a promising solution for mitigating factual errors by incorporating relevant external information.}

\begin{table*}[t]
    \centering
    \scriptsize
    \begin{tabular}{p{15.5cm}}
        \toprule
        \textbf{Triple: }(Chicago, country, United States of America) \hfill \textbf{Entity Popularity: $95.0\%ile$}\\
        \textbf{Question:} What country is Chicago located in? \hfill\textbf{Entity-Relation Popularity: $97.4\%ile$}\\ %
        \textbf{LM Answer:} United States {\color{darkgreen}[Correct]}\\
        \textbf{Context: } The Chicago Municipal Tuberculosis Sanitarium was located in Chicago, Illinois, USA.\dots{\color{darkgreen}[Correct Retrieval]}\\
        \textbf{RALM Answer:} USA {\color{darkgreen}[Correct]}\\\midrule
        \textbf{Triple: }(George H.W. Bush, educated at, Yale University) \hfill \textbf{Entity Popularity: $89.5\%ile$}\\
        \textbf{Question:} What educational institution did George H.W. Bush attend? \hfill \textbf{Entity-Relation Popularity: $41.8\%ile$}\\
        \textbf{LM Answer:} Yale University {\color{darkgreen}[Correct]}\\
        \textbf{Context: } The George H.W. Bush Presidential Library is located on a site on the west campus of Texas A\&M University in College Station, Texas.\dots {\color{crimson}[Wrong Retrieval]}\\
        \textbf{RALM Answer:} Texas A\&M University {\color{crimson}[Wrong]}\\\midrule
        \textbf{Triple: }(Ellen Litman, educated at, University of Pittsburgh) \hfill \textbf{Entity Popularity: $10.3\%ile$}\\
        \textbf{Question:} What educational institution was Ellen Litman educated at? \hfill \textbf{Entity-Relation Popularity: $17.9\%ile$} \\ %
        \textbf{LM Answer:} Stanford University {\color{crimson}[Wrong]} \\
        \textbf{Context: } Ellen Litman Ellen Litman (born 1973) is an American novelist. She received the Rona Jaffe Foundation Writers' Award in 2006. Born in Moscow, Russia, she emigrated with her parents in 1992 to Pittsburgh, Pennsylvania. She was educated at the University of Pittsburgh and earned a B.S. in Information Science. \dots{\color{darkgreen}[Correct Retrieval]}\\
        \textbf{RALM Answer:} University of Pittsburgh {\color{darkgreen}[Correct]}\\\bottomrule
    \end{tabular}
    \vspace{-2mm}
    \caption{QA examples from WiTQA with predictions of varying popularity of question entity and entity-relation pair. We show the predictions from LM (GPT-3.5) with no augmentation and RALM (GPT-3.5+BM25). In the top row, both LM and RALM provide correct answers for the popular question. In the middle row, LM generates correct answer but RALM provides incorrect answer due to retrieval errors. In the bottom row, LM provides incorrect answer for an infrequent entity-relation pair.
    }
    \vspace{-3mm}
    \label{tab:examples}    
\end{table*}

{Nevertheless, recent studies suggest that RALMs are not a universal solution~\cite{petroni2020how,li-etal-2023-large}. Indiscriminately augmenting LMs with irrelevant passages can override potentially correct knowledge already possessed by the LM, resulting in incorrect responses (illustrated in Table \ref{tab:examples}). A robust RALM is characterized by its ability to accurately recall its prior knowledge while selectively incorporating retrieved information only when necessary.}

Determining when to recall and when to retrieve external information thus requires a thorough investigation of the following questions:
 \begin{enumerate} %
\item Under what conditions LMs can recall correctly and what factors influence their ability?
\item When retrieval augmented models make errors and what factors affect their performance?
\item Are there any common error patterns between LMs and retrieval models responses?
\end{enumerate}
\noindent

\noindent
{While previous research has investigated factors influencing memorization in LMs as well as performance of retrievers, they have a few limitations:
a) They solely focus on entities ~\cite{sun2023headtotail,mallen-etal-2023-trust}, whereas real-world information comprises both entities and relations . b) They primarily focus exclusively on either retrievers or recall in LMs, neglecting the interplay between them ~\cite{petroni-etal-2019-language, sciavolino-etal-2021-simple,liu2023pre}.}

{
In this work, we aim to provide a finer-grained (fact-centric) analysis by investigating the impact of combinations of entities and relations on the performance of RALMs. We focus on question answering (QA) task, where we analyze 10 LMs of varying sizes} with 5 different retrieval settings.
{To facilitate this, we require a QA benchmark that not only provides valid supporting passages for each QA pair but also integrates indicators for  memorization in LMs. 
Additionally, the benchmark should encompass entities and relations of varying popularity. However, existing benchmarks such as PopQA ~\cite{mallen-etal-2023-trust} and EntityQuestions~\cite{sciavolino-etal-2021-simple} are not suitable for this purpose as they 
are entity-centric and target long-tailed information. To facilitate fact-centric analysis, we develop a novel dataset called \textsc{WiTQA} (Wikipedia Triple Question Answers). We sample triples extracted from Wikipedia, considering the popularity of entities and entity-relation combinations. We then generate QA pairs for each triple, ensuring that each example in \textsc{WiTQA} is accompanied by supporting passages and popularity scores for the question entity-relation pair.}

Our investigation of RALMs zero-shot performance on the proposed benchmark dataset (\textsc{WiTQA}) yields the following key findings:

\begin{itemize} %
    \item Even without retrieval, LMs can correctly recall entity-relation pairs frequently encountered during pre-training. Nonetheless, this capability is notably contingent on the model's size. Larger models can acquire long-tailed relations about renowned entities. However, there is still a significant drop in overall performance when addressing minor facts.
    \item For long-tailed entity-relation pairs, retrievers show more robust performance compared to recall abilities of LMs, suggesting that augmentation for such cases is beneficial.
    However, this observation does not extend to well-known entity-relation pairs, potentially leading to override issues.
    \item LMs achieve higher accuracy than retrievers for well-known entity-relation pairs regarding long-tailed entities while previous studies report that large LMs struggle with long-tailed entities. 
\end{itemize}

Our findings reveal when retrieval does not assist LMs through the lens of fact-level popularity.
Leveraging this insight, we propose a selective memory integration, which selectively employs retrieval augmentation and LMs' memory based on the frequencies of entities and relations. Our experiments demonstrate that this approach can enhance QA performance by up to $10.1$\%.

\begin{figure*}[t]
    \centering
    \includegraphics[width=.99\textwidth]{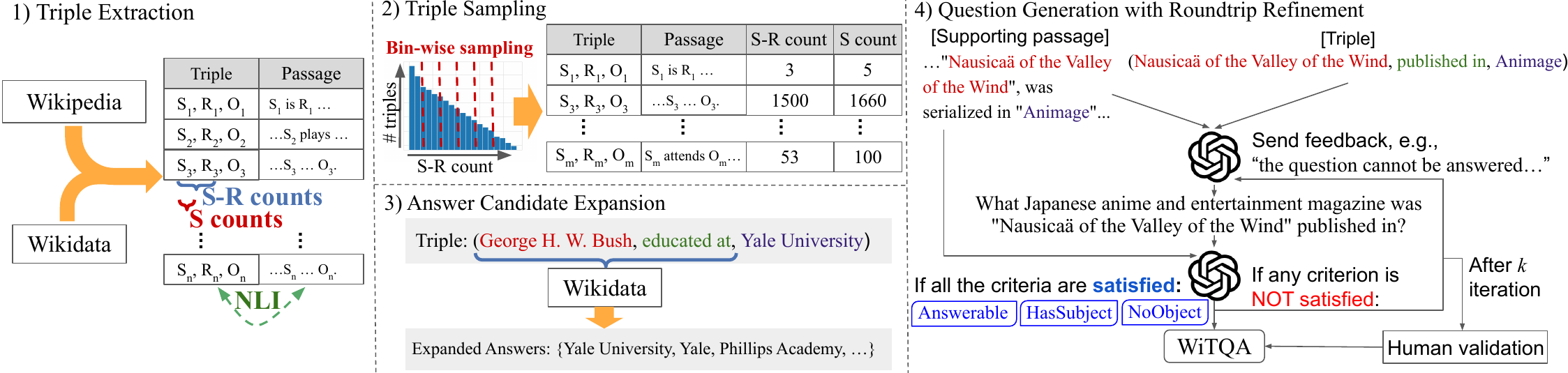}
    \vspace{-2mm}
    \caption{Overview of WiTQA dataset creation. First, we extract triples from Wikipedia and Wikidata, and compute the frequency of subject-relation pairs and subject entity (referred to as S-R counts and S counts)  (\S\ref{sub:triple_extraction}). Second, we sample triples based on different ranges of S-R counts and select supporting passages based on entailment scores(\S\ref{sub:triple_sampling}). Third, we expand answer candidates using  Wikidata (\S\ref{sub:answer_expansion}). Finally, we generate questions from triples and iteratively refine generated questions (\S\ref{sub:roundtrip}). }
    \vspace{-2mm}
    \label{fg:data_creation}
\end{figure*}

\section{Background}
\label{sec:background}
\subsection{Open domain Question Answering}
Open domain question answering is a knowledge-intensive task that involves generating an answer as an output $a$ given a question $q$ as an input. This task typically involves retrieving relevant passage $p$ based on the given question $q$ using a retrieval model $\mathcal{M}_{\texttt{ret}}$. Subsequently, the retrieved passage is used to model the answer $a$ using a reader model $\mathcal{M}_{\texttt{read}}$: $\mathcal{M}_{\texttt{read}} (a | q, p)$.

\subsection{Parametric vs Non-Parametric Knowledge}
Although, in the early study of the open domain question answering, a reader model $\mathcal{M}_{\texttt{read}}$ always relied on the external passage $p$ to reliably answer the question $q$~\cite{chen-etal-2017-reading,radford2019language,lee-etal-2019-latent,pmlr-v119-guu20a,lewis2020retrieval}, recent LM-based reader models started showing the potential to answer the question without relying on the external passage $p$: $\mathcal{M}_{\texttt{read}}(a | q, p = \phi)$~\cite{petroni-etal-2019-language,roberts-etal-2020-much,brown2020gpt3,mallen-etal-2023-trust,yu2023generate,pmlr-v202-kandpal23a,kang2023impact,shi2023large}.
This has actually shed light on the prospect that retrieval model $\mathcal{M}_{\texttt{ret}}$ can have a negative impact on RALMs. For example, \citet{li-etal-2023-large} showcased that if the irrelevant passage is provided to the LMs, the parametric knowledge in LM is overridden by the non-parametric knowledge in the passage $p$. In the context of open domain question answering tasks, this suggests that the LM-based reader model $\mathcal{M}_{\texttt{read}}$ may produce incorrect answer $\bar{a}$ because it can be misguided by the irrelevant passage $\bar{p}$ provided by the retrieval model $\mathcal{M}_{\texttt{ret}}$.

To address this issue, various approaches have been proposed~\cite{yu2023chain,mallen-etal-2023-trust,yoran2023making,asai2023self}. For instance, \citet{mallen-etal-2023-trust} reported that combining LMs and RALMs based on the entity popularity can yield improved QA accuracy.  
\citet{asai2023self} built a training dataset to determine when to engage in retrieval and to assess the relevance of retrieved passages using GPT-4 and then fine-tuned smaller-scale models to replicate similar behavior.

However, none of the above addresses the distinct strengths and weaknesses of LMs and retrieval models, nor have they clarified the causes of errors when these systems are interconnected. This primarily stems from the lack of datasets designed for such in-depth analysis. Thus, we introduce the \textsc{WiTQA} dataset, specifically created to analyze the error patterns of LMs and retrieval models, both independently and in tandem.

\begin{table*}[t]
    \centering\scalebox{0.7}{
    \begin{tabular}{l|ccccccc} \toprule
        Dataset & Page view & S count & S-R count & Supporting passages & \# of Relation Type & Question form  \\\midrule 
        EntityQuestions \cite{sciavolino-etal-2021-simple} & \xmark & \xmark & \xmark & \xmark & $24$ & Template \\
        PopQA \cite{mallen-etal-2023-trust} & \cmark & \xmark & \xmark & \xmark & $16$ & Template \\
        WiTQA (Ours) & \cmark & \cmark & \cmark & \cmark & $32$ & Model-assisted \\
        \bottomrule
    \end{tabular}
    }
    \vspace{-2mm}
    \caption{Summary of question-answering datasets. WiTQA includes question popularity indicators and valid supporting passages sourced from Wikipedia. It contains more diverse relation types. It employs a model-assisted approach for question generation, leading to a more versatile verbalization of triples. 
    }
\label{tb:dataset}
\end{table*}

\section{The \textsc{WiTQA} dataset}
\label{sec:dataset}

The \textsc{WiTQA} dataset is meticulously constructed with the underlying assumption that LMs are likely to recall facts frequently mentioned in their pre-training corpus~\cite{petroni-etal-2019-language,jiang-etal-2020-know}. We curate question-answer pairs annotated with the frequency of mentions of entities and their relationships within Wikipedia—a predominant source of pre-training for LMs—along with their supporting passages. This enables us to explore the correlation between popularity in the pre-training corpus and the performance of LMs and retrieval models individually. Furthermore, when operating in tandem, it facilitates the analysis of whether RALM's errors stem from LM's reasoning or primarily arise from retrieval errors.

\subsection{Dataset creation}
\label{sub:data_creation}

Our process for creating the dataset involves four key steps: 1) extraction of triples from Wikipedia, 2) sampling of triples, 3) expansion of answer candidates, and 4) generation of questions with round-trip refinement, as illustrated in Figure~\ref{fg:data_creation}.

\subsubsection{Triple extraction}
\label{sub:triple_extraction}
We first extract triples from Wikipedia to estimate the popularity of the subject entity (S count) and the co-occurrence of the subject entity and relation predicate (S-R count) in the Wikipedia corpus.\footnote{\url{https://archive.org/download/enwiki-20211020/enwiki-20211020-pages-articles-multistream.xml.bz2}}
However, building a scalable and robust information extraction system is a long-standing challenge~\cite{chia-etal-2022-relationprompt,kim23zett}. To address this, we opted to leverage the Wikipedia hyperlink and Wikidata for rule-based triple extraction \cite{elsahar-etal-2018-rex, huguet-cabot-navigli-2021-rebel-relation}. Specifically, we extract a list of entities from the Wikipedia abstract, map them to Wikidata ID using Wikimapper,\footnote{\url{https://github.com/jcklie/wikimapper}} and finally extract all the triples mentioned in the text by matching them with the Wikidata database.

Given a list of extracted triples, we compute S count for each unique subject entity in the list. Similarly, we compute S-R count for each unique subject entity-relation pair in the list. 
Additionally, for each passage-triple pair, we compute entailment score using NLI models.\footnote{\url{https://huggingface.co/cross-encoder/nli-deberta-v3-large}} For a given triple, the passage with the highest entailment score is designated as the supporting passage. Please refer to Appendix \ref{app:sec:nli} for more details.

\subsubsection{Triple sampling}
\label{sub:triple_sampling}

Concerning subject-relation (S-R) counts, the distribution of triples follows a long-tail pattern. 
To ensure dataset diversity, we employed sampling based on S-R counts. Specifically, we manually sampled $32$ relations and categorized triples into intervals such as $[1,5)$, $[5,10)$, $[10, 50)$, $[50,100)$, $[100, 500)$, $[500, 1000)$, and $1000+$.
Subsequently, we sampled up to $200$ triples for each relation within each interval.

\subsubsection{Answer candidate expansion}
\label{sub:answer_expansion}

Given that a question formulated with a subject S and relation R can have multiple valid answers, it is crucial to recognize that the object O in a sampled triple (S,R,O) might not be the only correct response to the question. 
To address this issue, we define acceptable answers for a question derived from a triple (S,R,O) as a set of entities E for which (S,R,E) exists in  Wikidata. These acceptable answers include aliases of object entities listed in Wikidata.

\begin{figure}[t]
    \centering
    \includegraphics[width=0.48\textwidth]{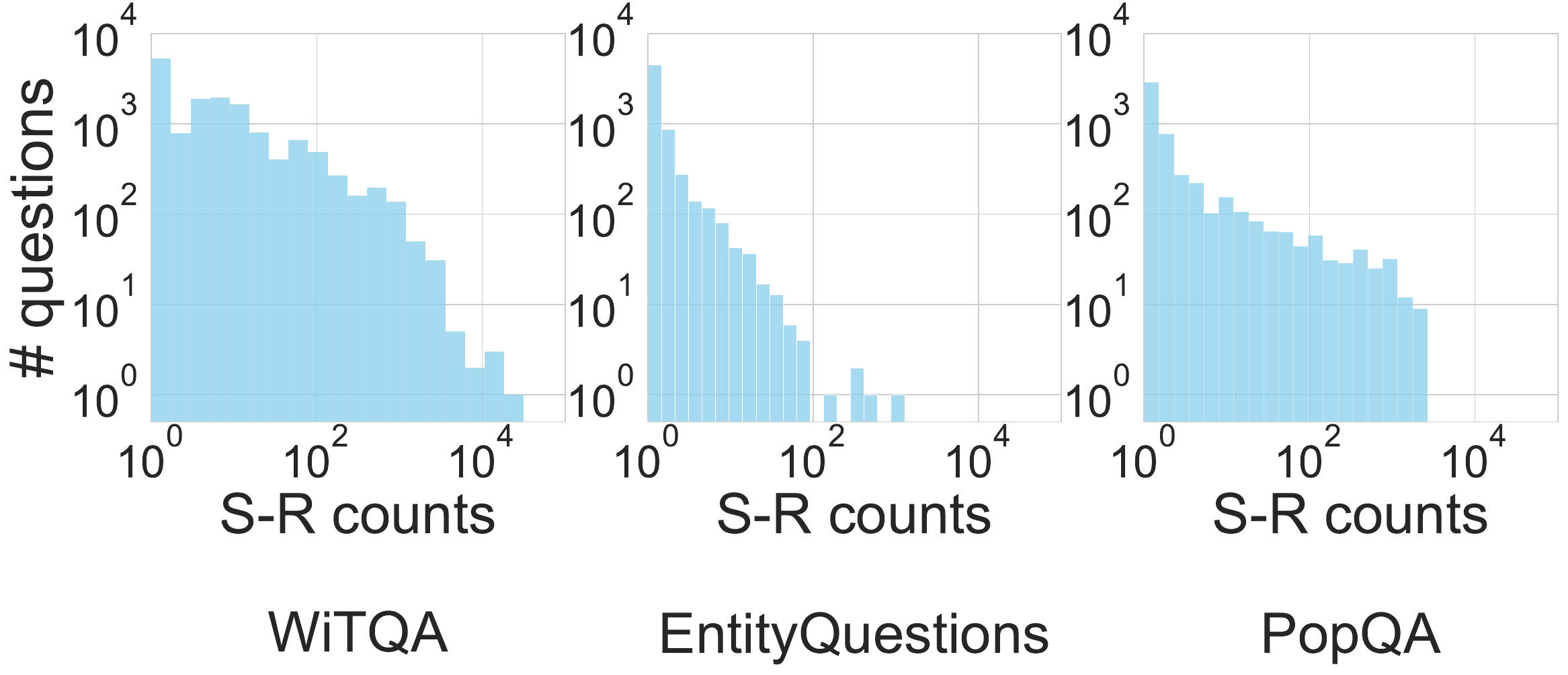}
    \caption{Histograms of question distributions. WiTQA exhibits greater diversity than existing benchmarks regarding question popularity, as indicated by the variation in S-R counts.}
    \label{fg:witqa_hist}
\end{figure}

\subsubsection{Question generation with roundtrip refinement.}
\label{sub:roundtrip}

Previous studies employ a template-based method to generate natural language questions for the triple-passage pairs. This method involves crafting a template for each relation with a placeholder for the subject entity, and treating the object entities as answers~\cite{sciavolino-etal-2021-simple,mallen-etal-2023-trust}. However, template-based approaches are known to suffer from questions of poor quality and diversity. For instance, consider the template \texttt{What sport does [SUBJ] play?} for the relation $\texttt{sport}$. This template works well for triples like $\texttt{(Shohei Ohtani, sport, baseball)}$, resulting in a natural question: \texttt{What sport does Shohei Ohtani play?}.  However, it becomes problematic with triples like \texttt{(2008–09 Maltese Premier League, sport, association football)}, leading to an awkward question: \texttt{What sport does the 2008–09 Maltese Premier League play?}.

One potential solution is to have each question written by a human; however, this approach is prohibitively expensive. Therefore, we adopt a model-assisted approach~\cite{zhang24coling} to automatically generate questions for each triple, based on the round-trip question generation method~\cite{alberti-etal-2019-synthetic}. Specifically, we generate a question given a context. We then use that generated question and the context to generate the answer. To assess the validity of a generated question, we establish three criteria:
\begin{itemize} %
    \item  \texttt{Answerable}: marked as True if the model accurately answers a generated question using its passage.
    \item  \texttt{HasSubject}: marked as True if a generated question includes the subject entity of its triple. 
    \item \texttt{NoObject}: marked as True if a generated question does not incorporate the object entity.
\end{itemize}
If a generated question fulfills all these criteria, it is considered valid; otherwise, feedback is provided to the model to enhance the question.
We use GPT-3.5 \cite{ouyang2022training} for this step due to its powerful language generation ability.

We observe that $95\%$ of questions ($14,094\allowbreak / 14,837$) satisfy all the criteria after three round-trip iterations. 
The remainder of the questions ($743 / 14,837$) were rewritten by our three internal annotators. We divided the datasets into three overlapping sections and consulted with at least two annotators to obtain human-written questions.
We used an annotation framework MEGAnno \cite{zhang-etal-2022-meganno} for this annotation due to its flexible labeling computational notebooks. 
Appendix \ref{app:sec:question_generation} shows all the prompts that we used for the question generation and verification. 

\begin{table}[t]
    
    \centering\scalebox{0.8 }{
    \begin{tabular}{l|r} \toprule
        Questions & $14,837$ \\ \midrule
        Unique subject entities & $13,251$ \\
        Unique object entities & $7,642$ \\ \midrule
        Average length of supporting passages (characters) & $214.3$ \\ \midrule
        Questions added in first roundtrip & $12,856$ \\ 
        Questions added in second roundtrip & $823$ \\ 
        Questions added in third roundtrip & $283$ \\ 
        Questions written by annotators & $743$\\
        \bottomrule
    \end{tabular}
    }
    \caption{Statistics of WiTQA.
    }
\label{tb:stats}
\end{table}

\subsection{Dataset Statistics}
\label{sub:witqa_stats} 
By following the outlined data creation process, the resulting dataset, \textsc{WiTQA}, comprises a total of $14,837$ questions.
In order to position \textsc{WiTQA} within the landscape of factoid QA benchmarks, we compare it with two established benchmarks: EntityQuestions \cite{sciavolino-etal-2021-simple} and PopQA \cite{mallen-etal-2023-trust}. Both of these benchmarks generate their questions from Wikidata triples. 
Table \ref{tb:dataset} illustrates that \textsc{WiTQA} is unique among QA benchmarks in that it includes supporting passages and question popularity. The histograms in Figure \ref{fg:witqa_hist} showcase the distribution of questions based on the S-R counts. Thanks to our bin-wise triple sampling, \textsc{WiTQA} features a substantial number of questions with over $1,000$ S-R counts, a characteristic that EntityQuestions and PopQA seldom possess.
Additionally, we observe that $62\%$ of questions in EntityQuestions and $65\%$ of questions in PopQA have never appeared in triples extracted from the Wikipedia abstract. In this context,  existing QA datasets have less diversity in terms of question popularity than \textsc{WiTQA}. 
For a more in-depth analysis of question popularity, we provide subject entity counts and subject page views\footnote{We obtain page views by querying the Wikipedia API.} for each question. 
We show the statistics of \textsc{WiTQA} in Table \ref{tb:stats}.
Detailed statistics of \textsc{WiTQA} are presented in the Appendix \ref{app:sub:additional_statistics}.

\section{Experiments: Recall or Retrieve}
\label{sec:eval}

We evaluate 10 language models of varying sizes augmented with four different retrieval methods to quantify the recall capability of LMs and the performance of retrievers in isolation and jointly and share the insights.

\subsection{Setup}

\noindent
\textbf{Models.} We use Flan-T5-small/base/large/xl (60M, 220M, 770M, and 3B) as small-scale LMs. We consider instruction fine-tuned Mistral-7B \cite{jiang2023mistral} and Llama-2-7B/13B/70B-chat \cite{touvron2023llama} for medium-scale LMs, and GPT-3.5/GPT-4 for large-scale LMs\cite{ouyang2022training,openai2023gpt4}.\footnote{gpt-3.5-turbo-0613 and gpt-4-0613}

\medskip

\noindent
\textbf{Retrievers.} We consider seven retrievers that include BM25~\cite{robertson2009probabilistic}, Contriever~\cite{izacard2021contriever}, GTR-large, GTR-xl~\cite{ni-etal-2022-large}, BGE\footnote{BAAI/bge-large-en}~\cite{xiao2023c}, GenRead~\cite{yu2023generate} and Oracle. 
BM25 is a static term-based sparse retriever that doesn't require training. Contriever is a unsupervised dense retriever pre-trained on a large corpora and fine-tuned on MS-MARCO~\cite{nguyen2016ms}. GTR and BGE are supervised dense retrievers pretrained on a large corpora and then fine-tuned on various supervised datasets including NQ \cite{kwiatkowski-etal-2019-natural} and HotpotQA \cite{yang-etal-2018-hotpotqa}. 
Both GTR and BGE retrieve from the  Wikipedia corpus that is also employed to create our benchmark. GenRead generates relevant passages by prompting large language models. Oracle retriever always retrieves the correct supporting passage for a given QA pair. We include it to measure the reasoning capabilities of the models in isolation from retriever errors. Please refer to Appendix \ref{app:retriever} for more details.
 
\medskip

\noindent
\textbf{Querying.} We use the following two templates for prompting a model.\footnote{\citet{mallen-etal-2023-trust} observed that more sophisticated instructions don't lead to significant improvements.} The first one (a) is generative prediction without any retrieval and (b) is contextual generative prediction that uses retrieved passage from one retrievers as a context.

\begin{center}
\begin{tikzpicture}
\node at (0,0) (a) [draw, rounded corners=4, fill=gray!20, align=left, text width=3.5cm] {Question:\{\{question\}\} \\ Answer:};
\node[left=-0.5cm of a, text width=4cm] {(a) Generative \\ Prediction:};
\node[below=0.5cm of a] (b) [draw, rounded corners=4, fill=gray!20, align=left, text width=3.5cm] {Context:\{\{context\}\} \\ Question:\{\{question\}\} \\ Answer:};
\node[left=-0.5cm of b, text width=4cm] {(b) Contextual \\ Generative Prediction:};
\end{tikzpicture}    
\end{center}

\noindent
\textbf{Metric.}
We mark a prediction as correct if any sub-string of the prediction is an exact match of any of the gold answers.

\subsection{Analysis of Model's Recall Ability}
\label{sub:recall}

\begin{figure}[t]
    \centering
    \includegraphics[width=0.48\textwidth]{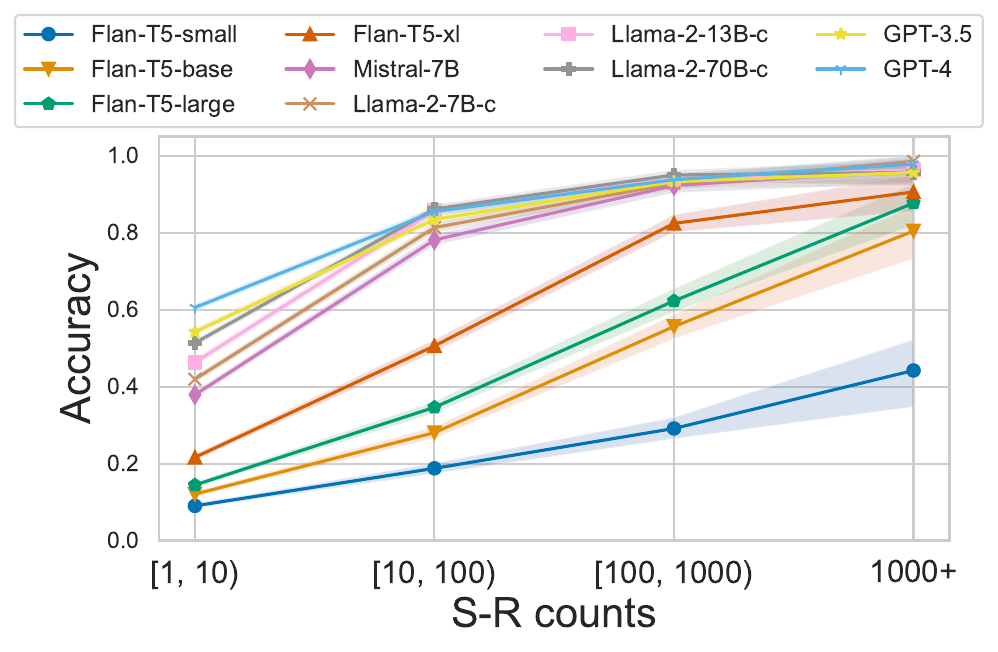}
    \vspace{-8mm}
    \caption{We categorize the questions into bins based on their S-R counts and present LMs accuracy across these bins.
    Shaded areas are the $95\%$ bootstrap confidence intervals with $1000$ samples. Larger models exhibit higher accuracy than smaller models. Even small models memorize factual knowledge about popular questions.}
    \label{fg:acc_sub_pred_models}
    \vspace{-1mm}
\end{figure}

\begin{figure}[t]
    \centering
    \includegraphics[width=0.48\textwidth]{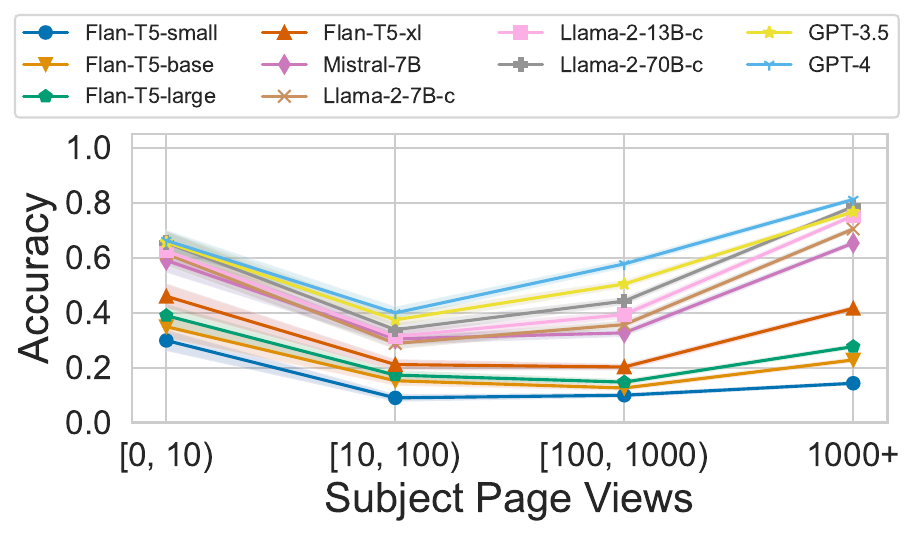}
    \vspace{-4mm}
    \caption{Accuracy over subject entity page views. }
    \label{fg:acc_page_view_models}
    \vspace{-4mm}
\end{figure}

\begin{figure}[t]
    \centering
    \includegraphics[width=0.48\textwidth]{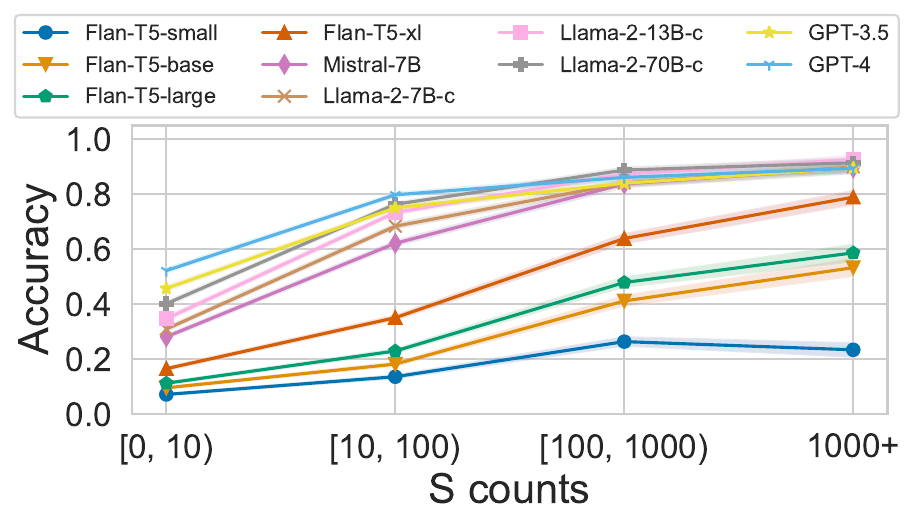}
    \vspace{-8mm}
    \caption{Accuracy over entity counts. }
    \label{fg:acc_ent_count_models}
    \vspace{-4mm}
\end{figure}

We explore the models' capacity to recall factual knowledge, considering models of various sizes and questions with differing popularity of subject-relation pairs. As depicted in Figure \ref{fg:acc_sub_pred_models}, all models, regardless of their size, generally demonstrate good recall of popular facts. For instance, even Flan-T5-large can achieve up to 80\% accuracy for questions with over 1000 S-R count. Predictably, larger models exhibit a superior ability to recall compared to smaller models.  Remarkably, in the case of less popular questions, there is a notable discrepancy in accuracy between small models and medium-/large-scale models. 
Some of these findings are consistent with the insights from recent works on evaluating factual knowledge in LMs \cite{sun2023headtotail}. 
Also, to compare S-R counts with subject page view counts which are used in the existing study \cite{mallen-etal-2023-trust} as the question popularity, we demonstrate that the accuracy of vanilla LMs does not consistently increase with increasing page view count in Figure \ref{fg:acc_page_view_models} while accuracy consistently increases as S-R and S counts increase in Figures \ref{fg:acc_sub_pred_models} and \ref{fg:acc_ent_count_models}, respectively. 
This suggests that our proposed metrics are more robust than previously proposed metrics.

\begin{figure*}[t]
    \centering
    \includegraphics[width=1.\textwidth]{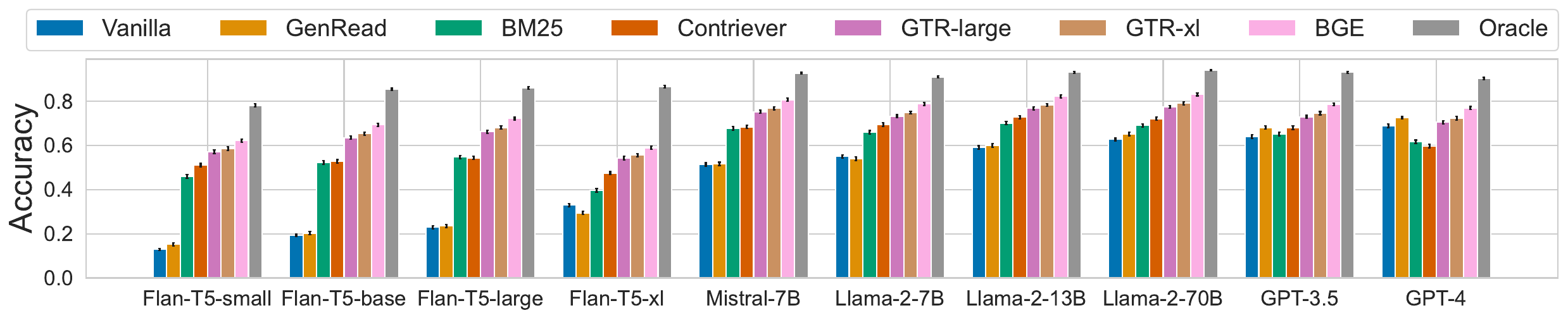}
    \vspace{-8mm}
    \caption{Overall accuracy of Vanilla LMs and LMs augmented with GenRead, BM25, Contriever, GTR-large, GTR-xl, BGE, and supporting passages (Oracle). Error bars show the $95\%$ bootstrap confidence intervals with $1000$ samples. Accuracy of Vanilla and GenRead improves with larger model sizes. 
    Augmentation with retrievers     enhances accuracy, especially for smaller models.  However, the gap diminishes and, in some cases, becomes negative for larger models such as the GPT series. Supporting passages (Oracle) prove beneficial for all models.
    }
    \vspace{-3mm}
    \label{fg:modelVSretrievers}
\end{figure*}

\subsection{When Do Retrievers Help}
\label{sub:ret}

Next, we augment the models with the retrievers and estimate the performances based on top-one retrieved passage. While expanding the context size could potentially enhance performance, we leave the exploration of augmentation with top-k passages for future investigations.

Figure \ref{fg:modelVSretrievers} shows the results of various LMs with and without augmentation.  Generally, retrieval augmentation enhances model performance, particularly for smaller models. However, the performance gap tends to be smaller, and sometimes even negative, for larger models like Llama-2 and the GPT series. GPT models often generate responses like ``The context does not provide information on...'' when the retrieved passages are insufficient to answer the question.We observe a similar pattern in Llama models, although it occurs less frequently. In essence, retrieved information often supersedes recall in larger models, suggesting that these models have high recall but are also more susceptible to retrieval errors.

Our observations indicate that BM25, Contriever, GTR, and BGE notably enhance the accuracy of small and medium models by up to $49.2\%$%
, while GenRead appears to be more effective for larger models with up to $4.1\%$ improvement. 
This suggests that the ability to maintain a coherent chain-of-thought emerges more prominently when the model size is sufficiently large.

We observe a significant improvement in accuracy for all models, irrespective of their sizes, when augmented with Oracle. Even the smallest model, Flan-T5-small, achieves an accuracy of $78.1\%$ with Oracle passages. However, substantial differences are observed between performances with BM25/Contriever/GTR/BGE and Oracle, which can be attributed to retrieval errors. When augmented with incorrect passages, most models struggle to provide correct answers.

Finally, we discuss why the small models augmented with Oracle passages obtain lower accuracy than larger models. We observe that when the context has multiple entities, the small models tend to predict the entity that appears close to the question words, which may not be correct. We give an example:
\begin{tcolorbox}%
Context:\\
Susanna Wesley (née Annesley; 20 January 1669 – 23 July 1742) was the daughter of Dr Samuel Annesley and Mary White, and the mother of John and Charles Wesley\\
Question:\\
Who was the mother of Charles Wesley?
\end{tcolorbox}
For example, Flan-T5-small answers “Mary White” while the true answer is “Susanna Wesley”. Note that the context supports the true answer, but Flan-T5-small cannot answer correctly. Highlighted by the example, we observe that small models tend to extract an entity as an answer, which appears near from the words in questions, leading to lower accuracy compared to larger models.

\begin{figure}
    \centering
    \includegraphics[width=0.48\textwidth]{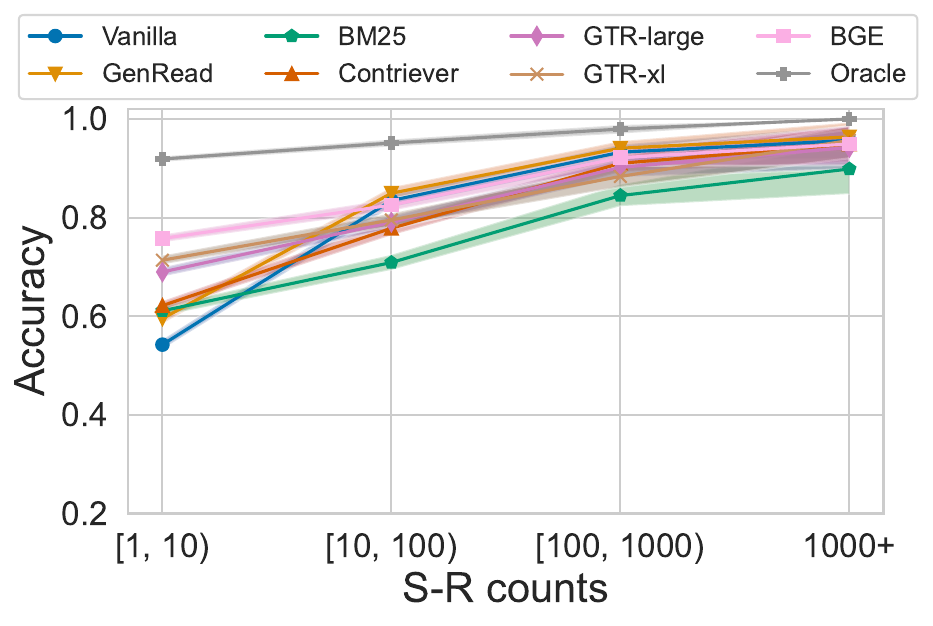}
    \vspace{-7mm}
    \caption{Accuracy of different retrievers with GPT-3.5 across varying question popularity.}
    \vspace{-2mm}
    \label{fg:pop_acc}
\end{figure}

\subsection{Deep Dive into Errors}
\label{sub:ralm_error}

In order to closely examine the factors influencing recall and retrieval, we concentrate on GPT-3.5 as the baseline model, along with the four retrievers.

\paragraph{How does popularity affect RALM performance?}

Figure \ref{fg:pop_acc} shows the accuracy of GPT-3.5 on questions with different subject-relation (S-R) counts. Notably, the accuracy of Vanilla and GenRead is considerable for popular questions but significantly declines when S-R counts drop below $10$. This drop is likely due to models memorizing popular facts. Conversely, RALMs utilizing BM25, Contriever, GTR, and BGE exhibit greater robustness on less popular questions. However, for popular questions, their performance is inferior compared to the Vanilla model without augmentation. Given that supporting passages (Oracle) consistently enhance the performance of models, we hypothesize that retrieval errors for popular questions negatively impact the performance of RALMs using BM25, Contriever, GTR, and BGE.

\begin{figure}[t]
    \centering
    \includegraphics[width=0.48\textwidth]{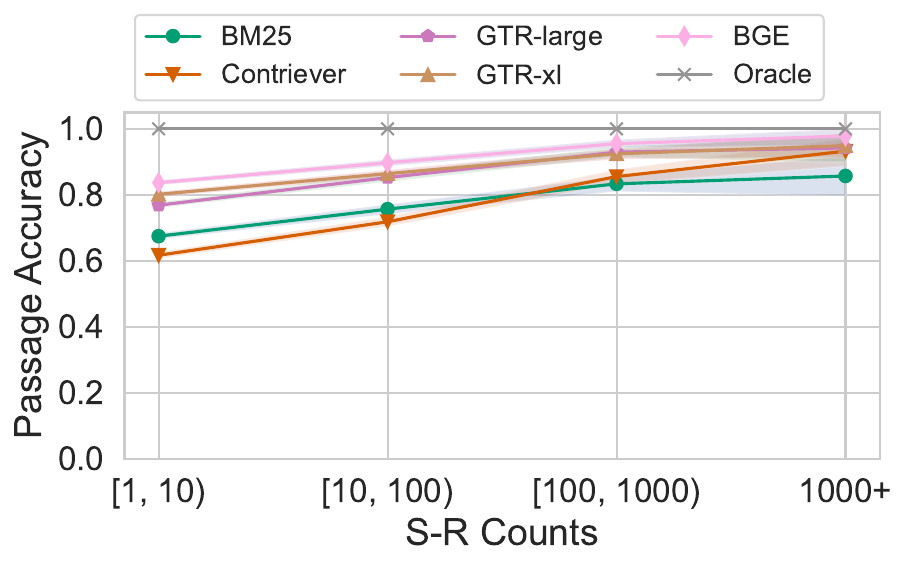}
    \vspace{-8mm}
    \caption{Passage accuracy across  question popularity.}
    \label{fg:retriever_acc}
    \vspace{-3mm}
\end{figure}

\paragraph{Is there a correlation between RALM performance and retrieval errors?}
To verify the above hypothesis, we closely examine the relationship between RALM performance and retrieval errors for questions of varying popularity. To estimate retriever performance, we compute passage accuracy, marking a passage as correct if any sub-string of the passage is an exact match for a gold answer.  We report the passage accuracy of different retrieval models in Figure \ref{fg:retriever_acc}. This trend aligns closely with RALM performance, as illustrated in Figure \ref{fg:pop_acc}. Specifically, 
agreement ratios between RALM performance and passage accuracy are $85.5\%, 82.4\%$, $88.5\%$, $88.4\%$, $88.7\%$, and $93.1\%$ for BM25, Contriever, GTR-large, GTR-xl, BGE, and Oracle, respectively.
Upon closer examination, we find that although retriever accuracy is low for rare questions, the drop is less significant compared to Vanilla LMs (see the leftmost plots in Figures \ref{fg:pop_acc} and \ref{fg:retriever_acc}). This indicates that retrieval augmentation is still beneficial for rare facts compared to recall. In contrast, Vanilla LMs perform better than retrievers for popular questions.

\begin{figure}[t]
    \centering
    \includegraphics[width=0.48\textwidth]{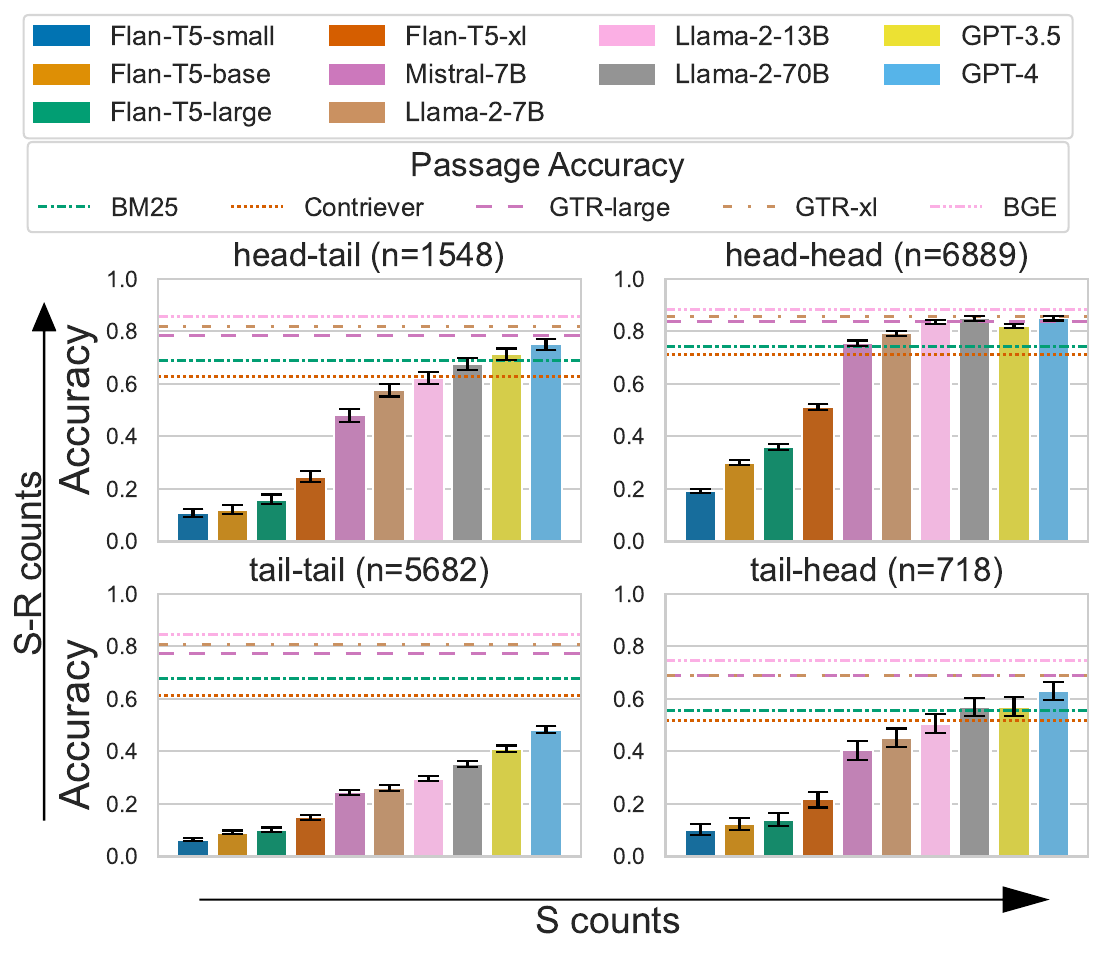}
    \vspace{-7mm}
    \caption{Analysis on Vanilla LMs with BM25, Contriver, GTR, and BGE passage accuracy over S-R counts and S counts ($n=$ the number of questions in the group). In the top row, S-R counts are higher than the median. In the bottom row, they are less than or equal to the median. In the left column, S counts are less than or equal to the median and in the right column, they are higher than the median.} 
    \label{fg:head_tail_vanilla}
    \vspace{-2mm}
\end{figure}

\paragraph{How does model size affect the need for augmentation?}
We further explore how the size of the model influences the need for augmentation based on the popularity of questions, considering that the ability to recall varies with model size. To this end, we categorize the questions into four groups based on the median of S-R counts ($5$) and S counts ($12$): 
\begin{itemize} %
\item head-head: both S-R and S counts are high
\item head-tail: S-R count is high but S count is low
\item tail-head: S-R count is low and S count is high
\item tail-tail: both counts are low
\end{itemize}

Figure \ref{fg:head_tail_vanilla} shows the accuracy of various models along with BM25, Contriver, GTR, and BGE passage accuracy
across the four groups. 
Note that we plot QA accuracy and passage accuracy since we use the same criteria for them. 
We observe that supervised retrievers, GTR and BGE, outperform vanilla LMs in all groups. This is attributed to the fact that it was fine-tuned with QA datasets about Wikipedia. For the generalizability to broader domains, we discuss the results with an unsupervised retriever BM25 in the following.\footnote{Similar results were observed for Contriever; hence, we focus on BM25 for simplicity.}

In the head-head group, medium and large models exhibit superior performance even without augmentation, e.g., the GPT series achieve higher than $80\%$ accuracy. In contrast, there is a notable gap between the accuracy of the vanilla model and passage accuracy for small models. This suggests that only small models need augmentation for questions in this group.

In the head-tail and tail-head groups, Llama and GPT series achieve comparable or superior accuracy to the passage accuracy. This indicates that large models can memorize long-tailed relations of popular entities and popular relations of infrequent entities. On the other hand, medium-scale models fail to recall compared to passage accuracy, and hence, they will benefit from augmentation. 
Interestingly, the passage accuracy in the tail-head group is significantly lower (less than $60\%$) compared to other groups. This retrieval error is caused by frequent mentions of the entity but infrequent mentions of subject-relation pairs. 
Consequently, accurately identifying relevant passages from a large pool of passages containing the question entity becomes challenging.

In the tail-tail group, even GPT series struggle to recall long-tail information for rare entities. Since retrievers are robust for such information, we conclude that retrieval augmentation is generally helpful for all models.

\paragraph{Does a combination of recall and retrieval improve accuracy?}
Based on these observations, we hypothesize that an optimal combination of recall and retrieval can be obtained using S-R and S counts as thresholds. We develop and examine a selective memory integration method that uses augmentation based on S-R and S counts, while using vanilla LM as default. This method yielded an improvement of 
$10.1\%$ and $8.1\%$
over Vanilla and BM25 for GPT-3.5 and GPT-4, respectively. For more details on this investigation, refer to the Appendix \ref{sec:method}.

\section{Further Related Work}
\label{sec:related}
Several recent studies \cite{asai2023self,yu2023chain,ma2023query,zhang2023merging} have focused on enhancing the robustness of RALMs by assessing the usefulness and relevance of retrieved passages to questions. 

\citet{yu2023chain} has proposed the Chain-of-Note framework, which systematically evaluates the relevance and accuracy of retrieved passages, thereby improving the noise robustness of RALMs.
\citet{ma2023query} has addressed query rewriting to improve retriever performance by using LMs or pre-trained LMs that are fine-tuned for a suitable query rewriter. 
\citet{zhang2023merging} has introduced an approach that leverages two sources of information: retrieved passages and LM-generated passages. 
This approach is designed on the hypothesis that answers corroborated by both sources have a higher likelihood of being accurate.

These studies rely on model-centric approaches for assessing document relevance to queries, refining queries, and integrating retrieved and generated passages. 
Thus, they overlook the importance of data-centric question popularity as an indicator for deciding when to retrieve and augment information. 
In contrast, our study leverages question popularity metrics derived from Wikipedia, offering insights that are complementary and distinct from these existing methodologies.

\section{Conclusion}
\label{sec:conclusion}

We introduced \textsc{WiTQA}, a novel QA dataset comprising of QA pairs associated with supporting passages. 
This dataset enables us to assess the performance of LMs with respect to question popularity and retrieval-augmentation. We conducted extensive experiments investigating the zero-shot performance of 10 LMs with four different retrieval methods.
Our findings reveal several key insights: 1) the ability to recall factual knowledge is influenced by the model's size, with even the GPT series facing challenges when addressing minor facts; 2) small RALMs demonstrate robust QA performance when provided with supporting passages, suggesting that errors in RALMs are primarily due to retrieval errors; and 3) retrievers exhibit greater robustness for long-tail information of long-tail entities compared to the recall capability of LMs.

\section*{Limitations}
\label{sec:limitation}
\paragraph{Distribution of pre-training corpus. }
This work hypothesizes that the distribution of the Wikipedia texts reflect that of pre-trained texts. 
However, the pre-trained texts of recent proprietary models such as GPT-4 are not accessible. 

\paragraph{Prompt engineering. }
A limited amount of prompt tuning was conducted. 
For example, larger models tend to refrain from generating answers when the provided passage does not pertain to the question.
From the viewpoint of maximizing QA accuracy, encouraging models to actively formulate answers regardless of passage relevance seems advantageous. 
Yet, this approach could elevate the proportion of incorrect answers, which is undesirable in practical applications. Exploring this trade-off will be a focus of our future work.

\paragraph{Multi-hop relations. }
Real-world questions often exhibit a complexity beyond simple triple-based questions, for instance, encompassing multi-hop relations. 
In this case, determining the subject and relation can be challenging. 
However, to conduct a deep analysis of LMs with clearly characterized questions, we focus on triple-based questions in this work. 

\section*{Ethical Consideration}
We use internal annotators for question rewriting in Section \ref{sub:roundtrip}, who were explained how data will be used by the authors directly and earned more than the minimal wage. 

\section*{Acknowledgment}
We thank Dan Zhang and Kushan Mitra for their help with annotation during the data creation process and Tanmay Laud and Estevam Hruschka for valuable discussion.

\bibliography{custom}

\appendix

\section{Detailed Setup for NLI Scores of Supporting Passages}
\label{app:sec:nli}

As discussed in Section \ref{sub:data_creation}, we use the entailment prediction to choose the best supporting passage for each triple. 
For each triple, we input the text containing both entities from the Wikipedia abstract, and the triple in their surface forms, subject + relation + object, separated by the <sep> token.
Then, we simply select the passage with the highest NLI score for each unique triple and use it as the supporting passage of the triple.

\section{Prompts for Question Generation/Refinement and Experiments}
\label{app:sec:question_generation}
\begin{figure}[ht]
\begin{tcolorbox}
Given a context and a triple (subject, relation, object), transform the triple to a question that asks ``Object''. The generated question must contain a given ``Subject'' and also be answerable without the context. \\
\\
\# Context:\\
\{\{ context \}\}\\
\\
\# Triple:\\
\#\# Subject:\\
\{\{ subject \}\}\\
\\
\#\# Relation:\\
\{\{ Relation \}\}\\
(Meaning: \{\{ relation\_description \}\})\\
\\
\#\# Object:\\
\{\{ object \}\}\\
\\
\# Output: <question only and must contain Subject>
\end{tcolorbox}
\caption{A prompt for question generation. ``\{\{ context \}\}'', ``\{\{ subject \}\}'', ``\{\{ relation \}\}'', and ``\{\{ object \}\}'' are replaced with the supporting passage, subject, relation, and object of a given triple. ``\{\{ relation\_description \}\}'' is a description of the relation, which is provided in Wikidata. }
\label{fg:app:qg_template}
\end{figure}

\begin{figure}[ht]
\begin{tcolorbox}
 \# Question:\\
 \{\{ question \}\}\\
\\
 \# Answer: <answer only> 
\end{tcolorbox}
\caption{A prompt template for vanilla. ``\{\{\ question \}\}'' is replaced with an actual question. }
\label{fg:app:qa_template}
\end{figure}

\begin{figure}[ht]
\begin{tcolorbox}
    Given a context and a question, answer the question.\\
    \\
    \# Context:\\
    \{\{ context \}\}\\
    \\
    \# Question:\\
    \{\{ question \}\}\\
    \\
    \# Answer: <answer only>
\end{tcolorbox}
\caption{A prompt template for GenRead, BM25, Contriever, and Oracle. ``\{\{\ question \}\}'' and ``\{\{ context \}\}'' are replaced with an actual question and context, respectively. }
\label{fg:app:rag_template}
\end{figure}

\begin{figure}[ht]
\begin{tcolorbox}
Generate a background document from Wikipedia to answer the given question. \{\{ question \}\}"
\end{tcolorbox}
\caption{A prompt template for GenRead passage generation. ``\{\{\ question \}\}'' is replaced with an actual question. This prompt comes from its original paper}
\label{fg:app:genread_passage}
\end{figure}

First, we provide a prompt used for our question generation in Figure \ref{fg:app:qg_template}. 
Next, we show prompts used for QA tasks in Figures \ref{fg:app:qa_template} and \ref{fg:app:rag_template}. 
Also, Figure \ref{fg:app:genread_passage} shows a prompt template to generate related passages by LMs for GenRead, which was introduced in its original paper. 

Also, Algorithm \ref{al:question_refinement} describes the roundtrip refinement. 
In line $1$, we initialize Message as Question generated by a prompt in Figure \ref{fg:app:qg_template}.
In line $3$, we obtain an answer to the question with its supporting passage by using a prompt in Figure \ref{fg:app:rag_template}.
Then, we check the three criteria ``Answerable'', ``HasSubject'', and ``NoObject'' for the question and answer pair. 
If the pair satisfies all the criteria, the question is added to WiTQA in line $6$. 
Otherwise we append a message that indicates which criteria are not satisfied in lines $8$--$27$.
In lines $17$--$20$, the algorithm describe an exceptional case in which the object of a triple is a substring of the subject and questions are answerable. Since questions cannot satisfy the criteria ``HasSubject'' and ``NoObject'' in this case, we use this condition to add such questions to WiTQA. 
Before proceeding to the next iteration, we regenerate a question with the message by using GPT-3.5 in line $28$. 
If we cannot obtain a question-answer pair satisfying the three criteria through $k$ iterations, the authors write the questions based on the triples. 

\begin{algorithm*}[ht]
    \small
    \caption{Roundtrip question refinement}
    \begin{algorithmic}[1]
        \Require Question, k
        \Ensure Message
        \State Message $\leftarrow$ [Question] \hfill $\vartriangleright$ Use a question generation prompt template in Figure \ref{fg:app:qg_template}
        \For{$i$ in range($k$)}
        \State Answer $\leftarrow$ GPT-3.5(Message) \hfill $\vartriangleright$ Use a RAG prompt template in Figure \ref{fg:app:rag_template}
        \State Answerable, HasSubject, NoObject $\leftarrow$ Check\_criteria(Question, Answer)
        \If{Answerable \& HasSubject \& NoObject}
            \State Add Question to WiTQA
            \State break
        \ElsIf{!Answerable \& HasSubject \& NoObject}
            \State Message.append(``It is good that the question contains `Subject' and not `Object', but the  
            \Statex\hspace{.95cm}question cannot be answered. Make the question more detailed if needed. Try again.'')
        \ElsIf{!HasSubject \& NoObject}
            \State Message.append(``The question you generated does not contain `Subject', but `Subject' must
            \Statex\hspace{.93cm} be in the question. Try again.'')
        \ElsIf{!HasSubject \& !NoObject}
            \State Message.append(``The question you generated does not contain `Subject' and does contain 
            \Statex\hspace{.97cm}`Object'. However, `Subject' must be in the question. Also, `Object' must not be in the 
            \Statex\hspace{1.02cm}question. Try again.'')
        \ElsIf{HasSubject \& !NoObject}
            \If{Question.subject in Question.object \& Question.subject $!=$ Question.object}
                \State Message.append(``The question you generated contains `Subject' and `Object', but 
                \Statex\hspace{1.35cm} `Object'   must not be in the question. Though `Subject' is the substring of `Object', 
                \Statex\hspace{1.48cm}remove `Object' and remain only `Subject'.'')
            \ElsIf{Question.object in Question.subject}
                \If{Answerable}
                    \State Add Question to WiTQA
                    \State break
                \Else
                    \State Message.append(``It is good that the question contains `Subject', but the question 
                    \Statex\hspace{1.95cm}cannot be answered. Make the question more detailed if needed. Try again.'')
                \EndIf
            \Else
                \State Message.append(``The question you generated contains `Subject' and `Object', but 
                \Statex\hspace{1.45cm}`Object' must not be in the question. Remove `Object' and remain only `Subject'.'')
            \EndIf
        \EndIf
        \State Message.append(GPT-3.5(Message)) \hfill $\vartriangleright$ Concatenate a refined question into Message
        \EndFor
    \end{algorithmic}
    \label{al:question_refinement}
\end{algorithm*}

\begin{table}[t]
\begin{tabular}{l|r}
\toprule
Relation label & Count \\
\midrule
    country & 1,269 \\
    sport & 1,034 \\
    capital & 1,032 \\
    capital of & 754 \\
    genre & 658 \\
    author & 654 \\
    language used & 639 \\
    country of citizenship & 619 \\
    father & 560 \\
    characters & 554 \\
    religion & 550 \\
    composer & 534 \\
    occupation & 518 \\
    publisher & 495 \\
    director & 479 \\
    place of birth & 443 \\
    educated at & 392 \\
    mother & 351 \\
    industry & 345 \\
    relative & 331 \\
    screenwriter & 278 \\
    producer & 271 \\
    doctoral advisor & 214 \\
    broadcast by & 214 \\
    published in & 210 \\
    location of first performance & 209 \\
    cuisine & 208 \\
    executive producer & 208 \\
    color & 207 \\
    medical condition & 204 \\
    architectural style & 203 \\
    director of photography & 200 \\
\bottomrule
\end{tabular}
    \caption{Number of questions containing relations in WiTQA. }
    \label{tb:predicate_count}
\end{table}

\begin{table}[ht]
    \begin{tabular}{l|r}
    \toprule
    Relation label & Count \\
    \midrule
    country & 1,979,521 \\
    sport & 613,742 \\
    country of citizenship & 278,973 \\
    place of birth & 210,125 \\
    genre & 169,649 \\
    capital & 164,597 \\
    occupation & 126,448 \\
    educated at & 90,799 \\
    director & 83,411 \\
    author & 75,810 \\
    capital of & 57,566 \\
    father & 43,401 \\
    publisher & 28,731 \\
    religion or worldview & 23,986 \\
    composer & 23,917 \\
    screenwriter & 16,754 \\
    producer & 16,578 \\
    language used & 15,296 \\
    mother & 10,348 \\
    industry & 10,118 \\
    characters & 9,783 \\
    architectural style & 7,428 \\
    relative & 4,327 \\
    doctoral advisor & 2,544 \\
    published in & 1,600 \\
    director of photography & 1,254 \\
    location of first performance & 1,086 \\
    broadcast by & 874 \\
    color & 787 \\
    medical condition & 771 \\
    executive producer & 707 \\
    cuisine & 417 \\
    \bottomrule
    \end{tabular}
    \caption{Relation counts in all extracted triples in Wikipedia abstracts. }
\label{tab:predicate_in_triples}
\end{table}

\section{Additional Experimental Details}
\label{app:sec:experiment}

\paragraph{Implementation. }
For open LMs, we execute all experiments on a GPU node with 8 NVIDIA A100-SXM cores. 
As for Llama-2-70B, we use AWQ quantization \cite{lin2023awq} to make it fit our GPU. 
We set the temperature parameter to $0$ for all experiments. 

\subsection{Additional WiTQA Statistics}
\label{app:sub:additional_statistics}
Table \ref{tb:predicate_count} shows $32$ relations used in WiTQA and the number of questions containing each relation in WiTQA. 

In Table \ref{tab:predicate_in_triples}, we show relation counts indicating how many times each relation appears in all extracted triples in Wikipedia abstract. 
Thanks to the bin-wise triple sampling described in Section \ref{sub:data_creation}, WiTQA successfully captures triples with long-tail relations such as ``cuisine'' and ``medical condition'', improving the diversity of its questions.

\begin{figure}[t]
    \centering
    \includegraphics[width=.48\textwidth]{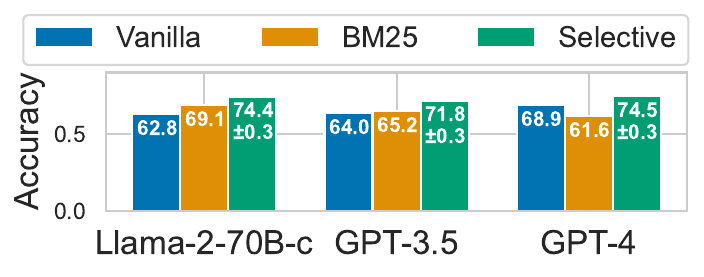}
    \vspace{-7mm}
    \caption{Accuracy of Vanilla, BM25, and LMs with a selective retriever. Accuracy ($\%$) is displayed inside bars. As for selective LMs, we run $5$ trials with different random seed to split a dataset and show the standard deviations. }
    \label{fg:selective}
    \vspace{-4mm}
\end{figure}

\begin{figure}
    \centering
    \includegraphics[width=.48\textwidth]{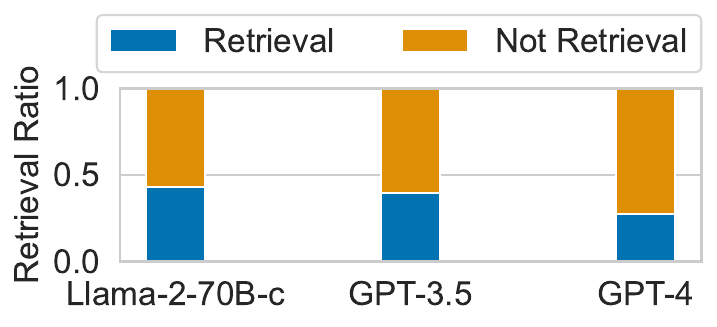}
    \vspace{-7mm}
    \caption{Average retrieval ratio of our selective memory integration method within $5$ trials. ``Retrieval'' represents the ratio of answers sourced from LMs with BM25, while ``Not Retrieval'' denotes the ratio of answers sourced from Vanilla LMs. }
    \label{fig:retrieval_ratio}
    \vspace{-2mm}
\end{figure}

\subsection{Selective Memory Integration}
\label{sec:method}
In Section \ref{sub:ralm_error}, our analysis revealed that while RALMs exhibit enhanced performance over Vanilla LMs for less popular questions, Vanilla LMs achieve superior performance in handling more popular questions. This finding suggests a complementary relationship between Vanilla LMs and RALMs, contingent upon the popularity of the questions. 
The above observation motivates us to selectively integrate LMs and RALMs with the larger sizes to improve overall accuracy by using the question popularity as an indicator to decide when to augment or not. 
As shown in Figure \ref{fg:head_tail_vanilla}, if both S-R and S counts are small (the tail-tail group), we need to augment LMs with retrieved passages; otherwise Vanilla and GenRead obtain higher accuracy. 
Hence, we take an approach that uses an RALM for the tail-tail group and Vanilla LM for other groups. 

\subsubsection{Settings}
\label{sub:setting}

We estimate optimal thresholds of S-R counts and S counts for each relation by using $50\%$ of questions in WiTQA, i.e., we find optimal thresholds for S-R and S counts so that the thresholds maximize the overall accuracy. 
Concretely, we use Vanilla LMs if questions with smaller S-R counts and smaller S counts than their thresholds, otherwise we use LMs with BM25. 
Then, we apply selective LMs accordingly to the remaining questions in WiTQA. 
We run the experiments with $5$ different random seeds to split the dataset and report the average and standard deviations. 

\subsubsection{Results}
\label{sub:results}
Figure \ref{fg:selective} shows that selective method achieves better accuracy than Vanilla and BM25 across all models. 
In particular for large models, the selective memory integration improves $7.7\%$, $10.1\%$, and $8.1\%$
over baselines for Llama-2-70B, GPT-3.5, and GPT-4, respectively. 
As discussed in Section \ref{sub:ret}, the GPT series with retrievers frequently output that the given contexts do not support the facts related to questions, rather than directly answering the questions. 
Consequently, their performance with selective integration is lower or comparable to that of Llama-2-70B.

Figure \ref{fig:retrieval_ratio} illustrates the average retrieval ratio of LMs with the selective memory integration over $5$ runs. 
We observe a trend that the retrieval ratio are shifting to smaller as the model size grows. 
For example, GPT-4 uses a retriever for only $27.7\%$ of questions while Llama-2-70B uses a retriever for $43.0\%$ of questions. 
This is mainly because larger models can typically recall facts more correctly than smaller models.

\subsection{Retriever setup}
\label{app:retriever}
\paragraph{Passage chunking}
We use the llama-index to chunk the Wikipedia documents into chunks with \texttt{chunk\_size = 256} and \texttt{chunk\_overlap = 0}.

\end{document}